# Diagnostic-Guided Longitudinal Modeling for Forecasting Retinal Atrophy Progression

Liyin Chen[1,4], Souvick Mukherjee[1], Ines Maria De Carvalho Lains[1], Nazlee Zebardast[1,2], Mengyu Wang[2], Tobias Elze[2], Jason I. Comander[1,3]

[1]Department of Ophthalmology, Massachusetts Eye and Ear, Harvard Medical School, Boston, MA, USA; [2]Harvard Ophthalmology AI Lab, Schepens Eye Research Institute, Harvard Medical School, Boston, MA, USA; [3]Ocular Genomics Institute, Massachusetts Eye and Ear, Harvard Medical School, Boston, MA, USA; [4]Broad Institute of MIT and Harvard, Cambridge, MA, USA

***Abstract*— Stochastic generative models are increasingly used for longitudinal imaging, but their added complexity may provide limited benefit when predictable disease-related change is small relative to technical variability. We treat model-class selection (stochastic vs deterministic) as an empirical step determined by a task-adaptive diagnostic. For a longitudinal image prediction task complicated by irregular follow-up, acquisition variability, and device heterogeneity, the diagnostic asks whether inter-visit image change is driven by time-dependent disease progression signals or time-independent acquisition variability. If a stochastic approach fails to yield useful predictive diversity, a more parsimonious, deterministic model class is selected. We applied this strategy to fundus autofluorescence (FAF) future image prediction. Applied to a heterogeneous Optos FAF archive (24,335 images from 9,708 eyes), the diagnostic indicated that global inter-visit FAF change was weakly associated with elapsed time, and source-anchored stochastic configurations produced little endpoint-resolvable sample variation. Guided by the diagnostic, we developed Temporal Retinal U-Net (TRU), a deterministic single-pass predictor conditioned on irregular imaging history and prediction horizon. Evaluated on a held-out cohort and two independent zero-shot transfer cohorts spanning rare-disease and cross-vendor shift, TRU achieved the strongest overall performance among the evaluated classical and deep-learning comparators on image-level and eye-specific progression measures, with lower precision in the smaller cross-vendor cohort. These findings support a diagnostic-guided strategy for longitudinal image prediction and demonstrate an example where deterministic predictions outperform stochastic models.**

## I. INTRODUCTION

Clinical imaging archives are heterogeneous information systems: visits occur at irregular intervals, image quality and acquisition conditions vary across sessions, disease labels are incomplete or mixed, and devices differ in field of view and intensity characteristics. A useful longitudinal prediction pipeline must therefore do more than generate visually plausible images. It must recover progression information that can be converted into eye-level monitoring endpoints, remain reliable under acquisition variability, and support reproducible inference at clinical scale.

For example, progressive retinal atrophy threatens central vision when hypoautofluorescent lesions expand toward the fovea, a pattern visible on fundus autofluorescence (FAF) imaging in geographic atrophy, Stargardt macular dystrophy, and other retinal pigment epithelium disorders [1], [2]. A clinically relevant question is not only whether atrophy is present at a single visit, but how an individual eye is changing over time and whether that change is moving toward the central (foveal) region. Longitudinal FAF imaging contains a patient-specific monitoring signal. Each eye contributes an irregular sequence of images acquired under real clinical variability, and the prediction task is to infer future spatial progression from that history.

Most computational work on retinal atrophy has focused on segmentation, lesion-area quantification, progression-rate estimation, or disease-conversion prediction [3], [4], [5], [6], [7], [8]. These approaches are clinically valuable, but they often compress spatial disease structure (e.g. by calculating the scalar area of the atrophy) before future monitoring endpoints are defined. Dense future-image prediction offers a complementary representation. By forecasting the future FAF image itself, the model preserves lesion geometry, foveal spatial relationships, and image context, allowing downstream measurements such as atrophy-area change, fovea-to-atrophy distance change, or spatial change-map similarity to be derived after prediction. This flexibility is important for patient-specific monitoring, where the most relevant endpoint may differ by disease stage, lesion location, or clinical research context.

Longitudinal medical-image prediction has increasingly adopted stochastic generative methods, including neural ordinary differential equations, conditional diffusion models, latent diffusion pipelines, and flow-matching approaches [9], [10], [11], [12], [13], [14]. These methods are attractive when the conditional future is genuinely multimodal or when sampling exposes clinically meaningful uncertainty. However, stochastic modeling should not be assumed to be necessary for every longitudinal imaging task. In slowly progressive retinal atrophy, the disease-relevant signal may be small relative to time-invariant anatomy, acquisition variability, registration residuals, and session-specific intensity differences. In that regime, sampling can add computational cost, validation complexity, and seed-dependent variability without producing clinically useful posterior diversity of predictions.

Alongside framework choice, the construction of training and inference inputs can influence patient-specific forecasting. Many generative prediction frameworks are trained by noising the target and do inference by sampling from pure noise or from some transformation of the most recent patient image. In situations where the disease progressing signal is slow-changing and subject to high technical noise, if the input distribution seen during training differs from the input distribution used during inference, the model may lack the inductive bias from training and sampling from the most recent image (we call this training-inference alignment) to solve the denoising problem for patient-specific longitudinal forecasting.

We therefore treat model complexity as a governed design choice and propose a task-adaptive diagnostic for testing whether stochastic sampling provides meaningful benefit for a specified longitudinal image prediction task. The diagnostic has two stages. First, an elapsed-time sensitivity screen measures how strongly global inter-visit change increased with follow-up interval. A small time-associated component was interpreted as a low progression-signal-to-technical-variability regime. Second, a posterior-concentration analysis evaluates whether source-anchored stochastic references produce meaningfully seed-to-seed predictive dispersion and endpoint gain deterministic regression. If raw inter-visit change has low useful time dependency, sample dispersion remained below the endpoint measurement resolution, and no endpoint gain is detected, a deterministic single-pass prediction would be selected as the parsimonious model choice.

Applied to a large, heterogeneous FAF archive acquired with Optos cameras, the diagnostic identified a low-signal regime in which global inter-visit variation was weakly time-associated and the evaluated source-anchored stochastic references produced limited endpoint-resolvable sample dispersion. Guided by this result, we developed Temporal Retinal U-Net (TRU), a deterministic predictor for irregular longitudinal FAF histories (Fig. 1). TRU uses the prior history image and continuous time-to-target information to predict a future FAF image. We evaluate on a patient-level hold-out from the development dataset and, without retraining or fine-tuning, on two independent zero-shot cohorts that introduce rare-disease shift and combined vendor-and-disease shift. Evaluated against classical references and three published deep-learning methods, TRU was the strongest overall on the evaluated metrics, with the cross-vendor cohort necessarily less conclusive owing to its smaller endpoint-eligible sample.

The contributions of this work are threefold. First, we formulate model-class selection for longitudinal retinal prediction as an empirical informatics diagnostic. Second, using controlled stochastic and deterministic configurations, we show that training-inference alignment is a dominant determinant of prediction quality in this task dominated by acquisition noise, and that aligned stochastic sampling does not produce a measurable advantage on evaluated metrics over deterministic prediction. Third, we introduce TRU as an openly released deterministic instantiation of this task that generalizes across disease and device shift. These results support selecting and implementing models in a way that is diagnostic-guided and aware of training-inference alignment, resulting in an efficient and effective approach to longitudinal disease progression prediction.

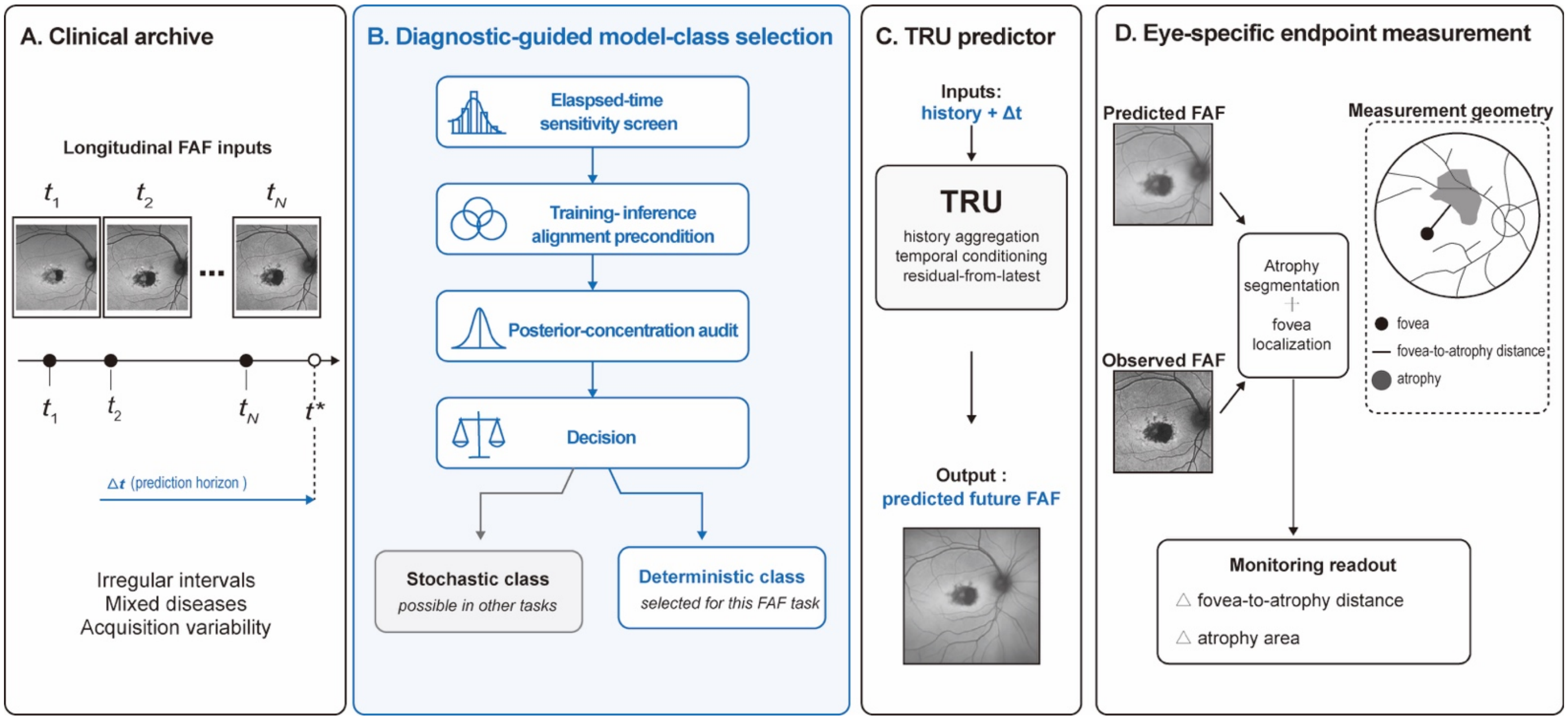

**Fig. 1.** Diagnostic-guided longitudinal FAF prediction pipeline.

## II. Methods

### A. Study Design, Cohorts, and Prediction Task

This retrospective study was approved by the Mass Eye and Ear / Mass General Brigham Human Studies Committee (Protocol 2025P001455) and was conducted in accordance with the Declaration of Helsinki. The requirement for informed consent was waived because the study used de-identified retrospective data.

The development cohort was an ultra-widefield FAF longitudinal dataset comprising 24,335 images from 9,708 eyes of 6,951 patients, acquired with a Optos ultra-widefield imaging platform, consisting of 2 generations of Optos camera acquisition technologies: OPTOMAP and OPTOMAP_PLUS. This dataset included mixed retinal pathology and eyes without pathology; diagnosis-code-identified Stargardt disease cases were excluded from the development set. Images were acquired during routine care with irregular follow-up intervals, with 2-10 FAF visits per eye and a median inter-visit interval of 0.51 years. A 90/10 patient-level split was used to define the development and held-out test cohorts, yielding 1,001 held-out Optos eyes for primary evaluation. An 85/15 patient-level split within the development cohort was used for training/validation.

Two independent zero-shot transfer cohorts were used to evaluate distributional shift relative to the development cohort. The first comprised 288 eyes from 195 patients with Stargardt disease with Optos imaging, to test rare-disease transfer within the same FAF device vendor family. The second comprised 122 eyes from 67 patients with Stargardt disease imaged with the Heidelberg Spectralis imaging platform, to test combined vendor and disease shift. No retraining, fine-tuning, or architecture modification was performed for either transfer cohort. Cohort characteristics are summarized in Table I.

The task is to predict the FAF image at a given future timepoint given the past FAF image(s) of the eye. For each eye, let $I_1, \dots, I_N$ denote a registered longitudinal image history acquired at times $t_1 < \cdots < t_N$, and let $I^*$ denote the target follow-up image acquired at time $t^* > t_N$ . The prediction task was to estimate $\hat{I}^*$ from the most recent image $I_N$, the available prior history $\{I_i\}_{i=1}^{N}$, and the elapsed time from each history frame to the target. For each frame, we defined $\Delta t_i = t^* - t_i$, with $\Delta t_N$ representing the prediction horizon from the most recent visit. All models were evaluated at the eye level, with patient-level clustering used for statistical inference.

## B. Task-adaptive Diagnostic for Model-class Selection

We formalized model-class selection as a prespecified decision procedure that selects the most parsimonious model-class choice from measured task structure and adopts stochastic generative modeling only where it demonstrably earns its added complexity. The thresholds in the diagnostic were selected based on the Optos development cohort (Algorithm 1). The procedure has three parts, a model-free screen, an alignment precondition, and a posterior-concentration audit. As every threshold is anchored to a quantity measured independently of the stochastic samples.

**Stage 1 (elapsed-time sensitivity screen).** On registered and intensity-harmonized inter-visit FAF pairs, and independent of any model, we computed for each pair the per-pixel absolute intensity changes and the changed-pixel fraction, defined as the fraction of pixels whose absolute intensity change exceeded 5 % on the [0, 1] normalized intensity scale. We then regressed the changed-pixel fraction on the elapsed interval Δt between the paired images and summarized the time-dependent fraction of inter-visit change $frac_{time}$,

$$frac_{time} = \frac{\beta_1 \, \Delta t_{med}}{\beta_0 + \beta_1 \, \Delta t_{med}}$$

where $\beta_1$ is the regression slope, the change in changed-pixel fraction per unit elapsed interval, $\beta_0$ is the intercept and gives the changed-pixel fraction extrapolated to Δt = 0, that is the time-independent component, and $\Delta t_{med}$ is the median elapsed interval across pairs. We quantified linear time dependence using Spearman correlation and the fitted slope from a regression of changed-pixel fraction on elapsed interval. Because progression-related change is expected to increase with elapsed interval, a small $frac_{time}$ relative to the intercept indicates that observed inter-visit change is dominated by time-independent technical variability, such as noise from acquisition, registration, and session effects. Given that the advantages of stochastic models mainly rely on the conditional posterior $p(I^* \mid I_N, \mathcal{H})$ having meaningful variance, a small $frac_{time}$ is flagged as a candidate for deterministic modeling. Stage 1 is used only as a candidate screen, and the binding decision will base on Stage 2.

**Alignment precondition.** A posterior-diversity probe is informative only for a model whose training and inference input distributions match, because a model that begins inference from a distribution it never saw in training may express its noise prior more than the conditional task. We therefore required the audited stochastic reference to be aligned, meaning that it was trained by corrupting the conditioning input and was run at inference from a corrupted version of the most recent observation. We showed that enforcing this alignment improves prediction for the given task (i.e. slow-progressing disease with high technical noise in images) (Section II.D and Section III.A).

**Stage 2 (posterior-concentration audit).** On the aligned reference we drew K = 10 samples per eye using different seeds, under a fixed history and fixed prediction horizon. Two criteria were assessed. The first criterion is resolvable diversity. For each eye, e, we computed the seed-to-seed standard deviation of the predicted endpoint $s_e$,

$$s_e = \sqrt{\frac{1}{K - 1} \sum_{k=1}^{K} (g_{e,k} - \bar{g}_e)^2}$$

where $g_{e,k}$ is the endpoint computed from the k-th sample of eye e, $\bar{g}_e$ is the mean of $g_{e,k}$ over the K samples, and K is the number of samples per eye.

**Algorithm 1** Task-adaptive model-class selection for longitudinal image prediction.

**Input:** longitudinal pairs $\{(H_i, I_i^*, \Delta_i)\}$ (history, target, horizon); aligned stochastic reference $R$; deterministic predictor, $D$; per-eye endpoint measurement-noise floor $\sigma_{\text{meas}}$ (Monte-Carlo module-error propagation, §2.5); prespecified negligible-effect margin $\delta \approx 0.064$ on the endpoint correlation; samples per eye, $K$

**Output:** selected class $C \in \{$STOCHASTIC, DETERMINISTIC$\}$ and audit trail

**Stage 1 — Elapsed-time sensitivity screen** *(model-free)*
$(m, c, \rho_{\text{time}}) \leftarrow$ REGRESS(changed-pixel fraction $\sim$ elapsed time)
$frac_{\text{time}} \leftarrow \dfrac{m \cdot \widetilde{\Delta t}}{c + m \cdot \widetilde{\Delta t}}$ *(time-attributable fraction at median horizon* $\widetilde{\Delta t}$*)*
**if** $frac_{\text{time}} \ll c$ **then**
  *screen* ← DETERMINISTIC CANDIDATE
**else**
  *screen* ← STOCHASTIC CANDIDATE
**end if**
  *indicative only; the binding decision is Stage 2*

**Precondition — Alignment**
**if** $R$ is not training–inference aligned **then**
  retrain $R$ aligned: corrupt the *conditioning* input, not the target; run inference from a corrupted latest observation
**end if**
  *diversity from a misaligned model reflects its noise prior, not the task*

**Stage 2 — Audit** $R$ *($K$ samples/eye, fixed conditioning)*
*(i) Resolvable diversity*
**for** each eye $e$ **do**
  $s_e \leftarrow \text{SD}_k\left[\text{endpoint}(R_{k,e})\right]$ *(seed SD of predicted* $\widehat{\Delta d}$*)*
**end for**
$resolvable \leftarrow \left(\text{median}_e\, s_e \geq \sigma_{\text{meas}}\right)$ *(else too few eyes clear the floor)*
*(ii) Endpoint gain*
$\Delta r \leftarrow \rho_S(R) - \rho_S(D)$ with patient-clustered 95% CI *($\rho_S$: endpoint correlation)*
$gain \leftarrow (\Delta r \geq \delta)$ **and** (CI lower bound $> 0$)

**Decision** *(parsimony: complexity must be earned)*
**if** *resolvable* **and** *gain* **then**
  **return** $C =$ STOCHASTIC $(R)$
**else**
  **return** $C =$ DETERMINISTIC $(D)$
**end if**

We deemed stochastic diversity resolvable only where $s_e$ reached the endpoint measurement-noise floor $\sigma_{\text{meas}}$, the per-eye standard deviation of the endpoint prediction as estimated by module-error propagation. In this study, the endpoint of fovea-to-atrophy distance was used; its standard deviation was influenced by errors in foveal localization and segmentation error (See section 2.5). A seed-to-seed standard deviation that stayed well below the measurement-noise floor across eyes indicated that sampling could not express clinically resolvable variation. We also summarized the mean pairwise inter-sample SSIM relative to the true disease-associated structural variation observed between consecutive real visits.

Stage 1 and both Stage-2 thresholds (σ_meas and δ) were derived on the Optos development cohort, whereas the alignment precondition and the Stage-2 audit statistics they adjudicate were computed on the Optos hold-out, avoiding any circular analysis that arises when a decision boundary is tuned on the same sample that supplies the statistic compared against it.

**Decision**. Stochastic prediction was selected only if the stage-2 criteria are met, and otherwise deterministic single-pass prediction was selected on parsimony grounds. The rule therefore places the burden of justifying added complexity on the stochastic class and refuses complexity that produces neither resolvable diversity nor an endpoint gain.

### C. Temporal Retinal U-Net

TRU is a deterministic single-pass predictor for irregular longitudinal FAF histories. TRU was used as the primary model because the diagnostic indicated limited useful stochastic diversity for this task.

TRU uses a four-level encoder-decoder U-Net backbone with channel widths 64, 128, 256, and 512, GroupNorm, SiLU activations [15], ResBlocks [16], and skip connections. The model contains approximately 22.8 million trainable parameters. Three components adapt the backbone to irregular longitudinal imaging histories. First, each time interval $\Delta t_i$ is encoded as a continuous 256-dimensional sinusoidal embedding using log-spaced frequencies, followed by a two-layer multilayer perceptron. This provides a continuous representation of prediction horizon and accommodates irregular clinical visit intervals. Second, a weight-sharing history encoder is used to process each history frame at three spatial scales. For each scale $s$, the encoder produces per-frame feature maps $h_{i,s}$ . These features are aggregated using delta-weighted attention,

$$\alpha_{i,s} = \frac{\exp\big(g_s(e_i)\big)}{\sum_{j=1}^{N} \exp\Big(g_s\big(e_j\big)\Big)}, \qquad H_s = \sum_{i=1}^{N} \alpha_{i,s}\, h_{i,s},$$

where $e_i$ is the time-delta embedding for frame $i$, $g_s(\cdot)$ is a learned scale-specific projection, and $H_s$ is the aggregated history representation at scale $s$.

Third, the aggregated history features are injected into the U-Net encoder at corresponding spatial scales along the learned feature-channel dimensions followed by convolution, normalization, and nonlinear refinement. TRU predicts residual change from the most recent observation:

$$\hat{I}^* = I_N + F_\theta(I_N, \{I_i, \Delta t_i\}_{i=1}^{N}),$$

where $F_\theta$ is the temporal U-Net output. The final layer was initialized near zero so that the initial model approximated the copy-last baseline before training.

TRU was trained using masked mean squared error,

$$\mathcal{L} = \frac{1}{B}\sum_{b=1}^{B} \frac{\sum_p M_{b,p}\left(\hat{I}_{b,p}^* - I_{b,p}^*\right)^2}{\sum_p M_{b,p}},$$

where $M$ is the validity mask and the per-image normalization gives each eye equal contribution independent of retinal image coverage. Training used AdamW, learning rate $1 \times 10^{-4}$, weight decay $1 \times 10^{-4}$, 1,000-step linear warmup, cosine decay, gradient clipping at 1.0, exponential moving average with decay 0.999, batch size 12, and 150 epochs. Random horizontal flipping and affine augmentation were applied jointly to all frames from the same longitudinal sequence to preserve spatial correspondence. During training, aggregated history features were randomly set to zero with probability 0.03 to regularize against over-reliance on history in low-history-count eyes. At inference, TRU uses one deterministic forward pass without sampling, denoising iterations, or post-hoc averaging.

### D. Training-inference Alignment Configurations

To test whether performance was determined primarily by stochastic framework choice or by alignment between training and inference input distributions, we evaluated controlled configuration variants that shared the TRU backbone, optimizer, training data, and hyperparameter schedule, differing mainly in how the model input was corrupted during training and initialized during inference.

Std-DDIM-50step, a conventional target-noising reference with 50-step DDIM initialized from pure Gaussian noise [17], [18]. This configuration represents maximum mismatch between the training input distribution and the patient-specific inference requirement.

Std-DDIM-1step used the same trained weights as Std-DDIM-50step but initialized inference from a noised version of the most recent patient image $I_N$, followed by a single-step prediction. This configuration isolates the effect of inference-side anatomical anchoring while leaving target-noised training unchanged.

Input-aligned-Nonlinear (IA-Nonlinear) used the same nonlinear diffusion noising schedule but noised $I_N$, not $I^*$, during training. Its inference input was also noised $I_N$, producing alignment between the training and inference input distributions.

Input-aligned-Linear (IA-Linear) used an aligned stochastic formulation with a linear interpolation schedule, constructing training inputs as $(1-t)I_N + t\epsilon$, with $t \sim U(0.05, 0.95)$, and using single-step inference from a noised version of $I_N$. This configuration tested whether the shape of the stochastic corruption schedule mattered once training and inference were aligned.

TRU represented the deterministic fully aligned configuration: it was trained and evaluated from the latest patient image and longitudinal history without stochastic noising. The alignment configurations are summarized in Fig. 2B.

### E. Personalized Progression Endpoints and Supportive Image Metrics

To demonstrate how image prediction can be used to derive clinically relevant information, the primary endpoint was change in fovea-to-atrophy distance. This was selected because it indexes whether atrophic lesions are moving toward the region most relevant to central vision. For each eligible eye, an automated pipeline was used to estimate the fovea coordinate in FAF images. Briefly, foveal coordinates were first detected from corresponding OCT B-scans using a model adapted from the open-source Choroidalyzer pipeline [19]; then, the coordinates were mapped to FAF by cross-registration between FAF and the SLO images co-acquired with OCT. Atrophy masks were derived from FAF images using an open-source segmentor trained with atrophic FAF image [9]. The minimum Euclidean distance from the foveal center to the atrophy boundary was denoted $d(I)$. Observed fovea-distance change was defined as

$$\Delta d = d(I^*) - d(I_N),$$

and predicted fovea-distance change was defined as

$$\Delta\hat{d} = d(\hat{I}^*) - d(I_N).$$

Agreement between predicted and observed change was summarized using Spearman correlation with patient-clustered 95% bootstrap confidence intervals for methods with nonconstant predicted change. For the copy-last baseline that produces constant zero change, rank correlation is undefined.

The secondary eye-level endpoint was change in total atrophy area. Let $a(I)$ denote the atrophy area measured from the segmentation mask. Observed and predicted atrophy-area changes were defined as

$$\Delta a = a(I^*) - a(I_N), \qquad \Delta\hat{a} = a(\hat{I}^*) - a(I_N).$$

Agreement was summarized using Spearman correlation with patient-clustered 95% bootstrap confidence intervals for nonconstant predictors. Endpoint analyses were restricted to eyes for which the corresponding fovea localization and atrophy segmentation measurements were well defined.

To validate the automated foveal localization underlying the fovea-to-atrophy distance endpoint, a retinal specialist manually annotated the foveal center for all eyes in the two Stargardt transfer cohorts. Automated foveal localization is most difficult in eyes with extensive macular atrophy, so these cohorts represented the most demanding condition for this step; manual annotation of the large mixed-disease cohort was not practical. Agreement between automated and manual foveal coordinates was quantified as the Euclidean distance between them on the 256 × 256 analysis frame and summarized by the median and interquartile range. As a sensitivity analysis, the fovea-to-atrophy distance change endpoint was recomputed using the manual coordinates, and the resulting method ranking was compared with that obtained from the automated coordinates.

To quantify how foveal-localization and atrophy-segmentation error propagate into the primary endpoint, we performed a Monte-Carlo perturbation analysis of B = 1,000 draws. On each draw, every foveal coordinate was displaced by a 2-D vector resampled from the empirical automated-versus-manual foveal residuals, and each atrophy boundary was perturbed by jittering the segmentation threshold ($\tau \sim N(0.5, 0.02)$), which shifts the boundary and hence the fovea-to-boundary distance; the resulting area perturbation (≈27 px²) was matched to the measured segmentation noise floor ($\sigma \approx 34$ px²). Identical perturbations were applied to the observed and predicted images, and the fovea-to-atrophy distance and its change were recomputed each draw. We used these draws for two purposes. First, as a prespecified sensitivity analysis (Section 2.7), we recomputed the endpoint correlation and the method ranking on each draw over the fixed endpoint-eligible eye set. Second, we quantified the endpoint measurement-noise floor, $\sigma_{meas}$, used in the Stage-2 resolvable-diversity criterion (Section 2.2). For each audit eye, $\sigma_{meas}$ was defined as the standard deviation of the recomputed fovea-to-atrophy distance change across 1,000 Monte Carlo draws. Among the development cohort, its distribution was right skewed (median, 5.17 px; IQR, 1.06-15.41 px). We therefore used the median of 5.17 px as the representative measurement-noise floor.

Supportive image-level metrics were used to assess whether predicted images preserved overall fidelity and progression structure. The principal image-level progression metric was change-map SSIM, denoted *Δ*SSIM, computed as the structural similarity between the observed change map $I^* - I_N$ and predicted change map $\hat{I}^* - I_N$ within the valid image region. PSNR was reported as a compact pixel-fidelity measure. Additional image metrics, including MAE and conventional SSIM, were computed for quality control but were not treated as primary evidence for patient-specific progression monitoring.

## F. Comparator Models

TRU was compared with five reference methods spanning persistence, linear extrapolation, continuous-time neural prediction, and stochastic generative prediction.

Copy-last used the most recent image $I_N$ as the predicted target and represented the no-progression clinical reference. Linear spline performed per-pixel linear extrapolation from the two most recent history frames to the target time. The deep-learning comparators were ImageFlowNet [9], TADM [11], and a 2D adaptation of BrLP [12]. ImageFlowNet and TADM were run from the authors' official public code (https://github.com/KrishnaswamyLab/ImageFlowNet; https://github.com/MattiaLitrico/TADM-Temporally-Aware-Diffusion-Model-for-Neurodegenerative-Progression-on-Brain-MRI), and 2D-BrLP was a faithful reimplementation following the published description (https://github.com/LemuelPuglisi/BrLP), adapting its 3D-brain latent-diffusion pipeline to single-channel 2D FAF. Each method's core mechanism was preserved, namely ImageFlowNet's multiscale neural-ODE trajectory formulation, TADM's temporally aware conditional diffusion, and 2D-BrLP's latent-diffusion progression in a 2D autoencoder latent space (approximately 247.9M parameters).

For fair comparison, every comparator used the identical preprocessing pipeline, train/validation split, and a matched training budget of 150 epochs with AdamW (weight decay 1e-4, 1,000-step warmup, cosine decay; effective batch size 12). Method-specific hyperparameters were then tuned on the development validation set. We searched the learning rate (TADM over {1e-4, 2e-4, 4e-4}; 2D-BrLP over {5e-5, 1e-4, 2e-4, 4e-4}; ImageFlowNet at its published 1e-4) and, for the diffusion comparators, the number of DDIM sampling steps (TADM over 2–50; 2D-BrLP over {10, 25, 50, 100}), selecting each model by its best validation SSIM and ImageFlowNet by its best validation PSNR. The selected configurations were TADM at learning rate 2e-4 with 5 DDIM steps and 2D-BrLP at learning rate 2e-4 with 50 DDIM steps and 5-sample latent averaging; higher learning rates destabilized TADM, with validation SSIM collapsing to 0.09 at 4e-4. Validation loss reached a plateau for every model, and each generative comparator was verified to produce non-degenerate, distributionally faithful samples. No held-out test or transfer cohort results were used for hyperparameter selection.

In addition to these image-generating comparators, we evaluated a non-generative, endpoint-specific reference that tests whether the image-based predictor improves on continuing each eye's own measured trend. Using the identical segmentation and foveal-localization pipeline, we derived the fovea-to-atrophy distance at every historical visit, fit a per-eye temporal trend, and extrapolated it to the target time t*. Two variants were used: a slope through the two most recent visits and an ordinary-least-squares fit over all available history. Because this reference requires multiple timepoints to calculate a slope, it is defined only for eyes with at least two prior visits.

Two additional longitudinal predictors (SADM [20] and a latent flow-matching model [14]) were considered but not included as primary benchmarks because their core assumptions were not compatible with this study design or they failed to reach nondegenerate validation performance under the same validated training budget. To avoid reporting under-trained baselines, we restricted the benchmark to methods that reached convergence under validated hyperparameters.

## G. Statistical Analysis and Stratified Robustness

All statistical comparisons were performed at the patient level. For paired comparisons between TRU and each comparator, we used a 2,000-resample patient-clustered paired bootstrap. Holm correction was applied within each cohort-metric combination across the planned baseline comparisons. Bootstrap p-values below $10^{-4}$ were reported at the bootstrap floor.

The Monte-Carlo error-propagation analysis was prespecified as a sensitivity analysis for the primary endpoint; perturbation bands are reported as 2.5-97.5th percentiles over 1,000 draws. From the same draws, we define the endpoint-gain margin, δ, as the measurement-noise-limited resolution of the endpoint correlation. For each of 500 pairs of independent perturbation draws we computed the Spearman correlation between the observed fovea-distance changes recovered on the two draws, and took $\rho_{xx}$ as the median over pairs, where $\rho_{xx}$ is the rank test–retest reliability of the observed endpoint under the measured foveal-localization and segmentation error. Classical attenuation then gives the endpoint correlation destroyed by measurement noise, $\delta = \frac{1}{\sqrt{\rho_{xx}}} - 1$, the smallest correlation gains not attributable to measurement error. On the Optos development cohort validation set, this gives $\rho_{xx} = 0.89$ and δ ≈ 0.064 (patient-clustered 95% CI 0.046-0.077).

For subgroup robustness, the Optos camera image hold-out cohort was stratified by accumulated monitoring history and by baseline atrophy severity. Monitoring-history strata were defined by increasing minimum numbers of available visits. Baseline severity strata were defined by tertiles of baseline atrophy burden among endpoint-eligible eyes. Within each stratum, fovea-distance-change agreement was summarized using patient-clustered bootstrap confidence intervals. These stratified analyses were prespecified as robustness analyses to assess whether the primary endpoint signal was concentrated in a narrow subset of eyes.

# III. Results

## A. The Diagnostic Selects Direct Regression for Slow-progressing FAF Prediction

To guide the choice of prediction framework, we first screened task structure with a model-free elapsed-time sensitivity screen of raw inter-visit image pairs (Fig. 2A). True disease progression should correlate with time. Across all training pairs in the Optos development dataset, we observed that the changed-pixel fraction was only weakly associated with elapsed interval. Specifically, the median absolute intensity change was 4.7%, where roughly 50% of valid pixels changed up to 5% between consecutive visits. However, a regression of changed-pixel fraction on elapsed time gave a dominant time-independent component (intercept 0.473) and a small slope, with the fitted time-associated fraction at the median interval being only 1.9% of inter-visit change ($\rho_{time}$ = 0.116 [0.095, 0.136]; 9,009 pairs). We performed this Stage 1 analysis on the Spectralis cohort as transfer-device consistency check. The Spectralis cohort showed a similarly weak association between changed-pixel fraction and elapsed interval (3.0%; $\rho_{time}$ = 0.138 [−0.003, 0.278]). This reproducibility across FAF from two different devices indicated that this task was dominated by interval-invariant technical noise over time-dependent disease progression signals.

Determining whether the task nonetheless benefits from stochastic structure requires probing a generative model, and that probe is meaningful only if the model is correctly specified. We therefore conducted an experiment using controlled configurations that shared the same backbone, training data, and optimization (Fig. 2B). A reference trained by corrupting the target image and sampled from pure Gaussian noise (Std-DDIM-50step) is maximally mismatched to patient-specific inference because its starting point contains no patient anatomy; this configuration fell below the copy-last reference on progression-sensitive metrics, including change-map SSIM (ΔSSIM, Fig. 2C). Anchoring inference to the patient's most recent image with weights unchanged (Std-DDIM-1step) increased ΔSSIM from 0.251 to 0.291 and PSNR from 21.74 to 22.77 dB on the Optos hold-out, isolating the effect of anatomical anchoring. Matching the training input distribution to that inference distribution by corrupting the latest image rather than the target (IA-Nonlinear) increased ΔSSIM further to 0.328 and PSNR to 23.26 dB. Similarly for the primary eye-level endpoint, the fully aligned configurations exceeded both misaligned references on fovea-distance-change agreement, by up to Δr = +0.155 [0.080, 0.229] ($p < 0.001$) relative to the single-step misaligned reference. Together, these experiments show that prediction quality was driven primarily by matching the corrupted-input distribution used for training to the distribution used to initialize inference.

Among aligned configurations, stochastic sampling conferred no advantage over deterministic regression on image fidelity or on the eye-level endpoint. IA-Nonlinear, IA-Linear, and the deterministic TRU were closely matched on the Optos hold-out (ΔSSIM 0.321 to 0.328; PSNR 23.13 to 23.26 dB), and on the primary endpoint the stochastic references were statistically indistinguishable ($P \gg 0.05$) from TRU, differing by Δr = +0.0015 for IA-Nonlinear and +0.0010 for IA-Linear on the 750 eyes with measurable baseline atrophy. TRU reached ΔSSIM 0.327 and PSNR 23.23 dB. The posterior-concentration audit explains this equivalence directly (Fig. 2D). Under fixed patient history and prediction horizon, independent samples from the aligned references were nearly identical, with mean inter-sample SSIM of 0.99985 and stochastic variance contributing 0.0148% of total prediction error for the nonlinear reference (0.99998 and 0.0021% for the linear reference). Propagated through the endpoint, this residual diversity moved the predicted fovea-to-atrophy distance change by a median of 0.19 px per eye, far below the 5.17 px measurement-noise floor, and it stayed at or below that floor across the distribution (90th percentile 34.1 versus 36.6 px), with 86% of eyes below the floor. The same collapse of prediction diversity held in the Spectralis cohort, where stochastic variance remained below 0.02% of prediction error. An endpoint gain indistinguishable from zero together with seed-level diversity below the measurement-noise floor are the two conditions under which the diagnostic selects deterministic prediction.

By the development-derived rule (Algorithm 1; Table III), neither the resolvable-dispersion criterion nor the endpoint-gain criterion was met in the Optos mixed-disease cohort. Direct regression was therefore selected, and that model-class choice was fixed before evaluation in the Stargardt transfer cohorts.

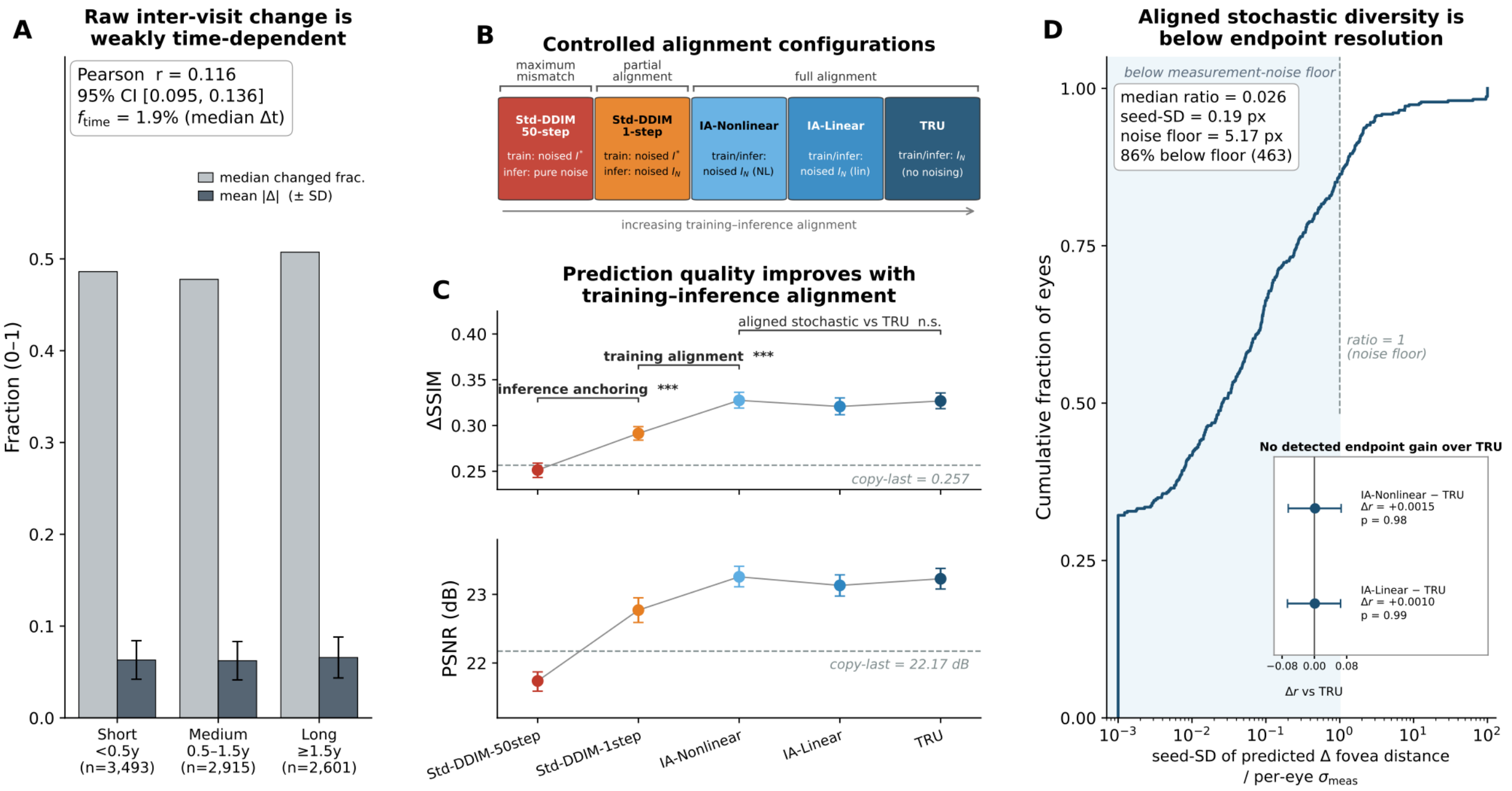

**Fig. 2.** Quantitative evidence for diagnostic-selected deterministic prediction. (A) Raw inter-visit FAF change was large but changed-pixel fraction was only weakly associated with elapsed interval across the 9,009 Optos training pairs, with similar changed-pixel fractions across short, medium, and long intervals and a small elapsed-time component at the median interval. (B) Controlled alignment configurations used to isolate inference anchoring, training-inference alignment, and deterministic single-pass prediction. (C) Prediction quality improved with inference anchoring and training alignment on change-map SSIM and PSNR, then plateaued among aligned stochastic and deterministic configurations. (D) In the aligned stochastic audit, seed-to-seed variability in predicted fovea-distance change remained below the per-eye measurement-noise floor for most eyes. In addition, aligned stochastic references showed no detected endpoint gain over TRU. Error bars denote 95% patient-clustered bootstrap intervals.

## B. TRU Recovered Patient-specific Fovea-threatening Progression

Table II summarizes the main benchmark across the Optos hold-out cohort and the two zero-shot Stargardt transfer cohorts. The primary endpoint was change in fovea-to-atrophy distance, because this clinically relevant endpoint measures whether atrophy is moving toward the high visual acuity, foveal region at the individual eye level. Change in total atrophy area was evaluated as a secondary endpoint, and ΔSSIM and PSNR were used as supportive image-level metrics.

On the Optos hold-out cohort, TRU achieved the strongest fovea-distance-change agreement among comparators, with Spearman correlation 0.50 and patient-clustered 95% CI [0.41, 0.58]. This exceeded the strongest comparator, TADM, by ΔSpearman = 0.12 (95% CI [0.03, 0.20]; patient-clustered paired bootstrap, Holm-adjusted $p = 0.006$). Linear spline showed substantially weaker endpoint agreement at 0.11 [0.02, 0.21]. Copy-last was retained as the no-progression reference; because it predicts zero endpoint change by construction, its rank correlation is not a valid endpoint-agreement statistic and is reported only as a persistence reference.

We next asked whether the performance of TRU reflected genuine spatial prediction or could instead be matched by continuing each eye's own scalar measured trajectory towards the fovea. On the eyes with at least two prior visits, where a per-eye trend is defined, extrapolating the patient's own fovea-distance history was anti-correlated with observed change (last-two-visit slope $r = -0.43$; OLS over all history $r = -0.37$), whereas TRU, given the identical imaging history, reached $r = 0.55$ on the same eye set (patient-clustered paired bootstrap, $p = 5\times10^{-4}$ for both variants). The negative correlation is consistent with regression to the mean in short, noisy histories: for most eligible eyes, only two prior visits were available, so the last measurement both determines the extrapolated slope and serves as the baseline for observed future change. Measurement error at that visit therefore induces an inverse association between the extrapolated and observed changes, most recent distance versus subsequent change, $\rho = -0.52$, whereas dense image-based predictors can exploit spatial lesion-boundary geometry.

TRU also had the highest atrophy-area-change correlation on the Optos hold-out cohort, with $r = 0.40$ [0.34, 0.46]. TADM was the closest comparator at 0.37 [0.30, 0.43], and the difference between TRU and TADM on this secondary endpoint was smaller than on the fovea-distance endpoint.

Supportive image-level metrics were consistent with the eye-level endpoint results. TRU achieved the highest ΔSSIM on the Optos hold-out cohort, 0.33 ± 0.14, compared with 0.30 ± 0.10 for 2D-BrLP, 0.29 ± 0.10 for ImageFlowNet, and 0.26 ± 0.13 for TADM. TRU also achieved the highest PSNR, 23.23 ± 2.38 dB, compared with 22.71 ± 2.54 dB for ImageFlowNet, 22.69 ± 2.70 dB for 2D-BrLP, and 22.29 ± 2.34 dB for TADM. The improvement over copy-last was most pronounced on ΔSSIM, indicating that TRU recovered spatial progression patterns beyond the most recent observation.

## F. Consistent Performance Under Disease and Vendor Shift

We next evaluated zero-shot transfer to Stargardt disease without retraining or fine-tuning. On the Optos Stargardt cohort, TRU retained the numerically highest fovea-distance-change correlation point estimate, r = 0.41 [0.28, 0.53]. TADM was the closest comparator at 0.34 [0.21, 0.48], while ImageFlowNet and 2D-BrLP both had point estimates of 0.28. TRU also had the numerically highest atrophy-area-change correlation point estimate, r = 0.34 [0.22, 0.46], compared with 0.20 [0.07, 0.32] for TADM and 0.20 [0.07, 0.31] for ImageFlowNet. On supportive image-level metrics, TRU again achieved the best point estimates on the Optos Stargardt cohort, with ΔSSIM of 0.35 ± 0.15 and PSNR of 23.66 ± 2.41 dB. The strongest image-level comparator was 2D-BrLP for ΔSSIM, at 0.33 ± 0.11, and ImageFlowNet for PSNR, at 23.12 ± 2.48 dB.

**Table I.** Dataset characteristics.

| Dataset | Purpose | Modality | Disease | Patients (N) | Eyes (N) | Images (N) | Split Strategy | Zero-Shot? |
|---|---|---|---|---|---|---|---|---|
| Optos Mixed-Disease | Primary Training & Eval | FAF | RPE pathologies, healthy | 6,951 | 9,708 | 24,335 | 90/10 Patient-level | No |
| Optos Stargardt | Rare-Disease Transfer | FAF | Stargardt dystrophy | 195 | 288 | 699 | No split | Yes |
| Spectralis Stargardt | Cross-Device Transfer | FAF | Stargardt dystrophy | 67 | 122 | 352 | No split | Yes |

**Table II.** Cross-cohort benchmark of TRU against published longitudinal medical image prediction methods and classical references. Three cohorts are organized by increasing distribution shift relative to the FAF mixed-disease training set: (A) in-distribution Optos FAF hold-out; (B) zero-shot Optos Stargardt (disease shift); (C) zero-shot Heidelberg Spectralis Stargardt (vendor + disease shift). All deep learning methods were trained exclusively on the Optos FAF mixed-disease training set with no identified Stargardt cases. Arrows indicate metric direction (↑ higher is better). Values are mean ± SD across eyes. Best value per metric per cohort is shown in bold. Cells are shaded within each cohort–metric column using a sequential colormap, with greener shading indicating better performance in the direction of the arrow in the column header; shading is normalized to the value range within each cohort and is not comparable across cohorts. * Indicates statistical significance ($P < 0.01$) between the best and the next-best methods. PSNR, in dB, is peak signal-to-noise ratio between predicted and observed target images; ΔSSIM is SSIM between the predicted and observed change maps, not a difference of SSIM values; r_growth is Spearman correlation between predicted and observed change in total atrophy area; and r_fovea_dist is Spearman correlation between predicted and observed change in minimum fovea-to-atrophy-boundary distance. PSNR and ΔSSIM are mean ± SD across eyes; correlation values are reported with patient-clustered 95% bootstrap CIs.

| Method | PSNR (dB) ↑ | ΔSSIM ↑ | r_growth [95% CI] ↑ | r_fovea_dist [95% CI] ↑ |
|---|---|---|---|---|
| ***A. Optos Holdout (N = 1,001 eyes)*** | | | | |
| Copy-Last | 22.17 ± 2.74 | 0.26 ± 0.11 | na | na |
| Linear Spline | 20.41 ± 4.47 | 0.20 ± 0.15 | 0.09 [0.02, 0.16] | 0.11 [0.02, 0.21] |
| TADM | 22.29 ± 2.34 | 0.26 ± 0.13 | 0.37 [0.30, 0.43] | 0.38 [0.29, 0.48] |
| ImageFlowNet | 22.71 ± 2.54 | 0.29 ± 0.10 | 0.23 [0.17, 0.30] | 0.32 [0.23, 0.41] |
| 2D-BrLP | 22.69 ± 2.70 | 0.30 ± 0.10 | 0.21 [0.14, 0.27] | 0.31 [0.21, 0.40] |
| TRU (This study) | 23.23 ± 2.38* | 0.33 ± 0.14* | 0.40 [0.34, 0.46] | 0.50 [0.41, 0.58]* |
| ***B. Optos Stargardt (N = 288 eyes) Within-vendor rare disease transfer*** | | | | |
| Copy-Last | 22.53 ± 2.91 | 0.29 ± 0.13 | na | na |
| Linear Spline | 21.29 ± 3.69 | 0.23 ± 0.16 | -0.03 [-0.14, 0.09] | 0.08 [-0.08, 0.25] |
| TADM | 22.59 ± 2.41 | 0.27 ± 0.13 | 0.20 [0.07, 0.32] | 0.34 [0.21, 0.48] |
| ImageFlowNet | 23.12 ± 2.48 | 0.31 ± 0.09 | 0.20 [0.07, 0.31] | 0.28 [0.14, 0.43] |
| 2D-BrLP | 22.97 ± 2.89 | 0.33 ± 0.11 | 0.15 [0.04, 0.26] | 0.28 [0.12, 0.42] |
| TRU (This study) | 23.66 ± 2.41* | 0.35 ± 0.15* | 0.34 [0.22, 0.46] | 0.41 [0.28, 0.53] |
| ***C. Spectralis Stargardt (N = 122 eyes) Cross-vendor/disease*** | | | | |
| Copy-Last | 20.01 ± 3.77 | 0.24 ± 0.11 | na | na |
| Linear Spline | 17.88 ± 4.67 | 0.15 ± 0.16 | 0.24 [0.02, 0.45] | 0.32 [0.03, 0.57] |
| TADM | 20.30 ± 2.02 | 0.19 ± 0.12 | 0.39 [0.20, 0.56] | 0.38 [-0.03, 0.68] |
| ImageFlowNet | 20.39 ± 3.68 | 0.26 ± 0.09 | 0.35 [0.14, 0.54] | 0.40 [0.03, 0.69] |
| 2D-BrLP | 20.10 ± 3.64 | 0.24 ± 0.08 | 0.36 [0.18, 0.52] | 0.47 [0.11, 0.73] |
| TRU (This study) | 21.72 ± 2.45* | 0.27 ± 0.15 | 0.45 [0.25, 0.61] | 0.51 [0.17, 0.77] |

**Table III.** Application of the selection rule (Algorithm 1) to the Optos longitudinal FAF task. Each criterion's statistic, observed value, externally anchored threshold, and resulting branch are shown. The aligned stochastic reference is IA-Nonlinear. Endpoint = fovea-distance-change Spearman correlation.

| Stage / criterion | Statistic | Observed | Threshold (anchor) | Branch |
|---|---|---|---|---|
| 1: Elapsed-time sensitivity screen | time-dependent fraction $frac_{time}$ ($\rho_{time}$) | 1.9% (ρ = 0.116 [0.095, 0.136]) | $frac_{time}$ ≪ component non-explained by time (intercept ≈ 0.47) | low time-associated global change → deterministic candidate |
| Precondition | endpoint effect of enforcing alignment | aligned configs exceed both misaligned references; +0.155 [0.080, 0.229] vs Std-DDIM-1step (p < 0.001) | audit only an aligned reference | aligned reference built |
| 2(i): Resolvable diversity | per-eye seed-SD of predicted distance-change vs σ_meas; inter-sample SSIM | median 0.19 vs 5.2 px; 86% sub-floor (median ratio 0.026); SSIM 0.99985 | seed-SD ≥ σ_meas (MC floor, §2.5) | fails (sub-floor) |
| 2(ii): Endpoint gain | aligned-stochastic − deterministic Δr [95% CI] | +0.0015 [−0.065, +0.067] (IA-Nonlinear, p = 0.98) | gain ≥ δ ≈ 0.064, CI lower bound > 0 | fails (no gain) |
| Decision | N/A | neither 2(i) nor 2(ii) met | stochastic only if both met | **deterministic (TRU) selected** |

The Spectralis Stargardt cohort tested combined disease and vendor shift. TRU again achieved the highest PSNR, 21.72 ± 2.45 dB, and the highest ΔSSIM point estimate, 0.27 ± 0.15. ImageFlowNet had a similar ΔSSIM point estimate of 0.26 ± 0.09, while 2D-BrLP and copy-last were both 0.24. TRU also had the highest fovea-distance-change point estimate, r = 0.51 [0.17, 0.77], followed by 2D-BrLP at 0.47 [0.11, 0.73]. TRU also had the highest atrophy-area-change point estimate, r = 0.45 [0.25, 0.61], compared with 0.39 [0.20, 0.56] for TADM, 0.36 [0.18, 0.52] for 2D-BrLP, and 0.35 [0.14, 0.54] for ImageFlowNet. However, confidence intervals were wide because the endpoint-eligible subset was smaller (N=56). Image-level results showed the same pattern. The Spectralis evaluation supports that TRU's performance was preserved despite vendor and field-of-view shift, though conclusions are necessarily limited by smaller sample size. Qualitative representative examples shows that TRU's predicted change maps reproduced the location and magnitude of the observed inter-visit change most closely, particularly the darkening at expanding lesion boundaries, whereas the generative comparators recovered the change pattern less completely (Fig. 3). Fig. 4 provides a complementary view of the same three eyes, showing the pixel-wise prediction error against the ground-truth follow-up image. The comparison makes the shared baseline failure mode visible, where every deep learning baseline and both reference methods leave a ring of positive residual at the advancing lesion boundary, indicating that they underpredicted the extent of atrophy expansion between visits. TRU's residuals at these boundaries are both spatially tighter and lower in magnitude, consistent with its quantitative advantage.

### G. Robustness of Fovea Endpoint Results

To confirm that the primary endpoint results were not due to artifacts in the fovea coordinate estimation, we compared the automated foveal localization with manual expert annotation in the two Stargardt transfer cohorts. The median Euclidean distance between automated and manual foveal coordinates was 4.55 px on the 256 × 256 frame in the Optos Stargardt cohort (IQR 2.35–8.24 px) and 18.03 px in the Spectralis Stargardt cohort (IQR 9.31–31.74 px). Recomputing the fovea-to-atrophy distance change endpoint with the manual coordinates did not alter the relative ranking of TRU in either cohort. Because automated localization agreed with expert annotation in the Optos Stargardt cohort (i.e the most atrophic eyes acquired on the Optos platform) and the Optos cohorts comprise predominantly less atrophic eyes, for which automated localization is more reliable, these results support the automated foveal localization used for the primary endpoint.

To further demonstrate that our reported results are robust to measurement errors, we propagated the measured foveal-localization and atrophy-segmentation errors into the endpoint by Monte-Carlo perturbation (Section 2.5). On the in-distribution Optos hold-out, the endpoint and the method ranking were kept unchanged. Across 1,000 perturbations matched to the measured localizer error, the fovea-distance-change correlation had a median of 0.48 with a 95% perturbation band of [0.44, 0.54] (excluding zero), and TRU remained the top-ranked nonconstant method in 100% of draws. TRU's performance advantage on the fovea metric remained significantly positive under disease shift (Optos Stargardt median 0.42, band [0.35, 0.48]) and under the substantially larger localization error of the cross-vendor Spectralis cohort (median 0.44, band [0.23, 0.63]).

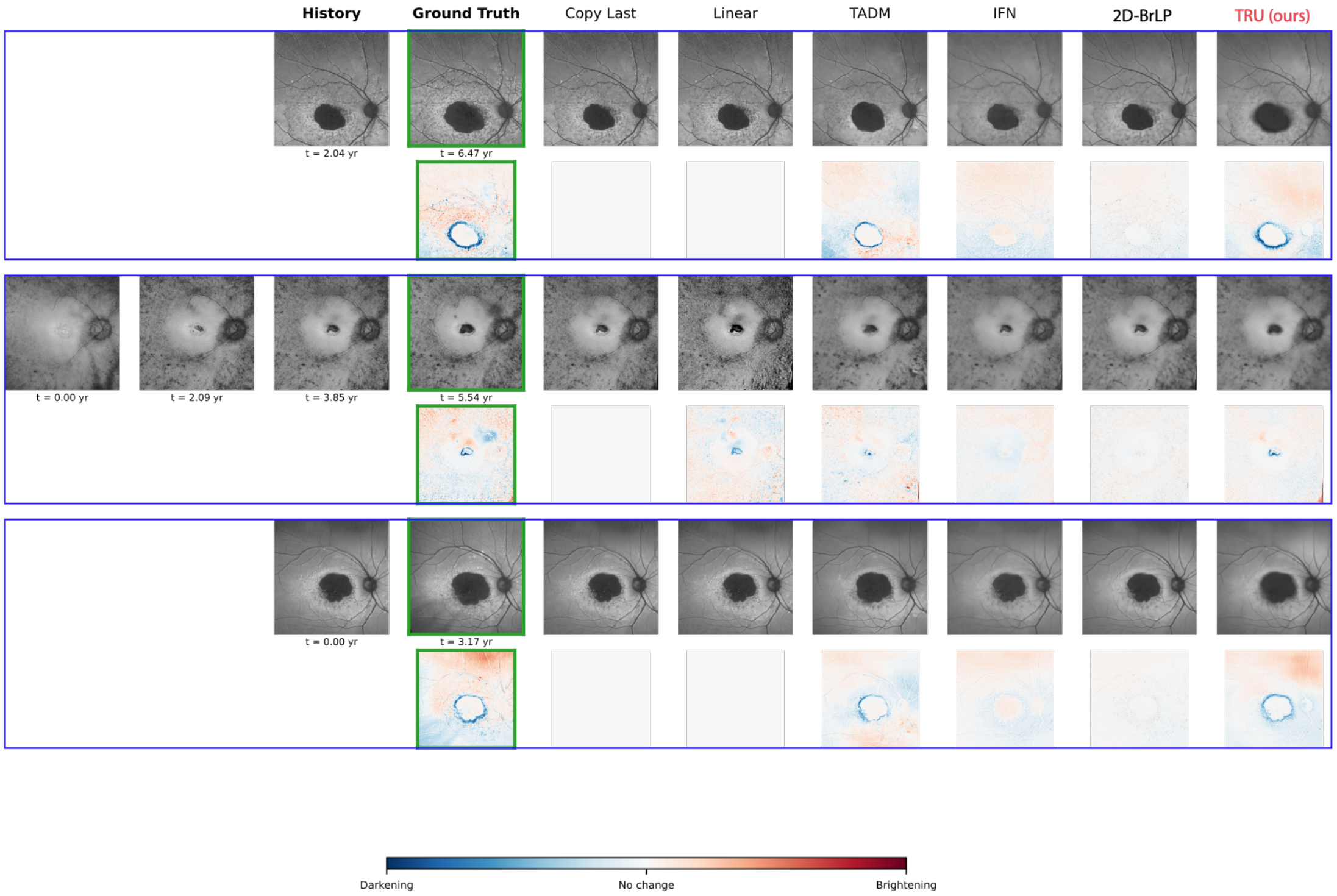

**Fig. 3.** Qualitative prediction examples on the Stargardt hold-out set demonstrate the zero-shot model performance. Each panel (blue box) contains one representative eye. The top row in each panel shows the longitudinal history frames, ground-truth target, and predictions from all evaluated methods. The bottom row in each panel shows the change maps, which display the pixel-wise difference between the predicted image and the most recent history frame, with blue indicating darkening (new atrophy) and red indicating brightening relative to the last observation. The ground-truth change map (green border) serves as the reference. Copy-last produces a blank change map by construction. Among deep learning methods, TRU's change maps most closely reproduce the spatial pattern and magnitude of the ground-truth change signal, particularly the localized darkening at lesion expansion boundaries.

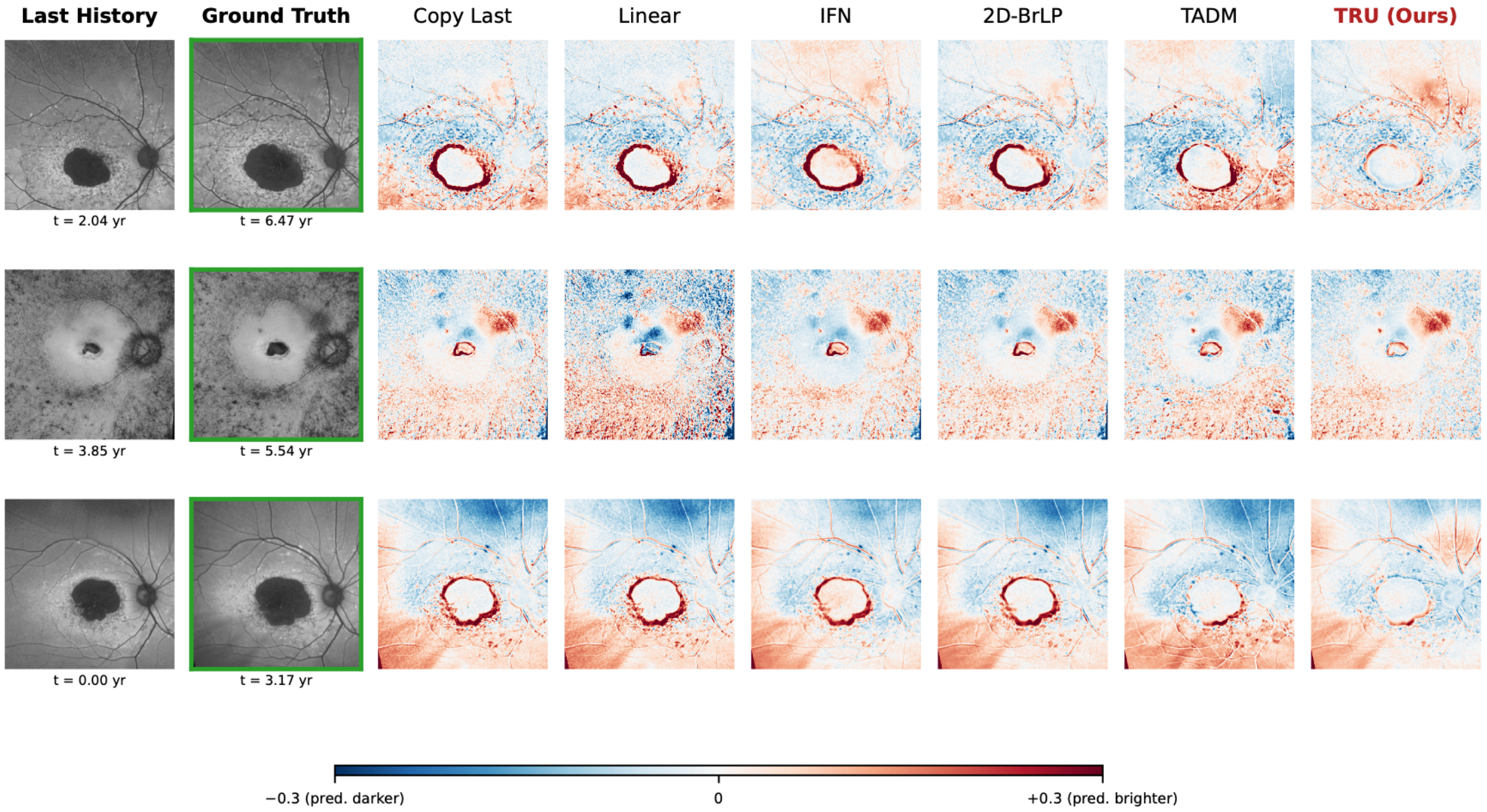

**Fig. 4.** Pixel-wise prediction error maps for three representative Stargardt eyes (same cases shown in Fig. 3). Each row corresponds to one eye; inter-visit timing is indicated beneath each frame. The first column shows the most recent history frame, and the second column (green border) shows the ground-truth target image I*. Direct side-by-side comparison of these two columns reveals the lesion expansion each method is asked to predict. Subsequent columns show the signed per-pixel prediction error (prediction − ground truth) for each evaluated method, rendered with a diverging colormap (blue: prediction darker than ground truth; red: prediction brighter than ground truth; symmetric scale ±0.3). Smaller-magnitude and spatially diffuse residuals indicate more accurate predictions. A characteristic ring of positive residual (red) at the atrophy boundary indicates that a method underpredicted the darkening associated with lesion growth between visits. This underprediction ring is prominent for copy-last, linear spline, ImageFlowNet, 2D-BrLP, and TADM across all three eyes, and is visibly attenuated for TRU, consistent with the quantitative results reported in Table II.

## H. Fovea-distance Prediction Was Robust Across History Length and Baseline Severity Strata

Finally, we evaluated whether the primary endpoint, the change in the fovea-to-atrophy distance (delta-fovea-distance), was concentrated in a narrow subset of Optos hold-out eyes or persisted across clinically-relevant heterogeneity (Fig. 5). When the hold-out cohort was stratified by accumulated monitoring history, TRU remained the highest-performing method across increasing minimum numbers of available visits. The point estimates did not show monotonic degradation as the history length increased, whereas comparator methods remained lower across the same strata. Because stricter history thresholds reduced the number of eligible eyes and widened bootstrap intervals, these results should be interpreted as robustness evidence, not as proof that additional visits necessarily improve prediction for every patient.

TRU also retained the highest correlations for delta-fovea-distance estimates across baseline atrophy-severity strata. The strongest point estimate occurred in the intermediate baseline-severity tertile, but better endpoint agreement with TRU was preserved in both earlier and more advanced baseline disease. The stratified analysis indicates that the primary endpoint result was not driven by a single baseline-severity band and that TRU preserved eye-level progression signal across the disease-burden range represented in the Optos hold-out cohort.

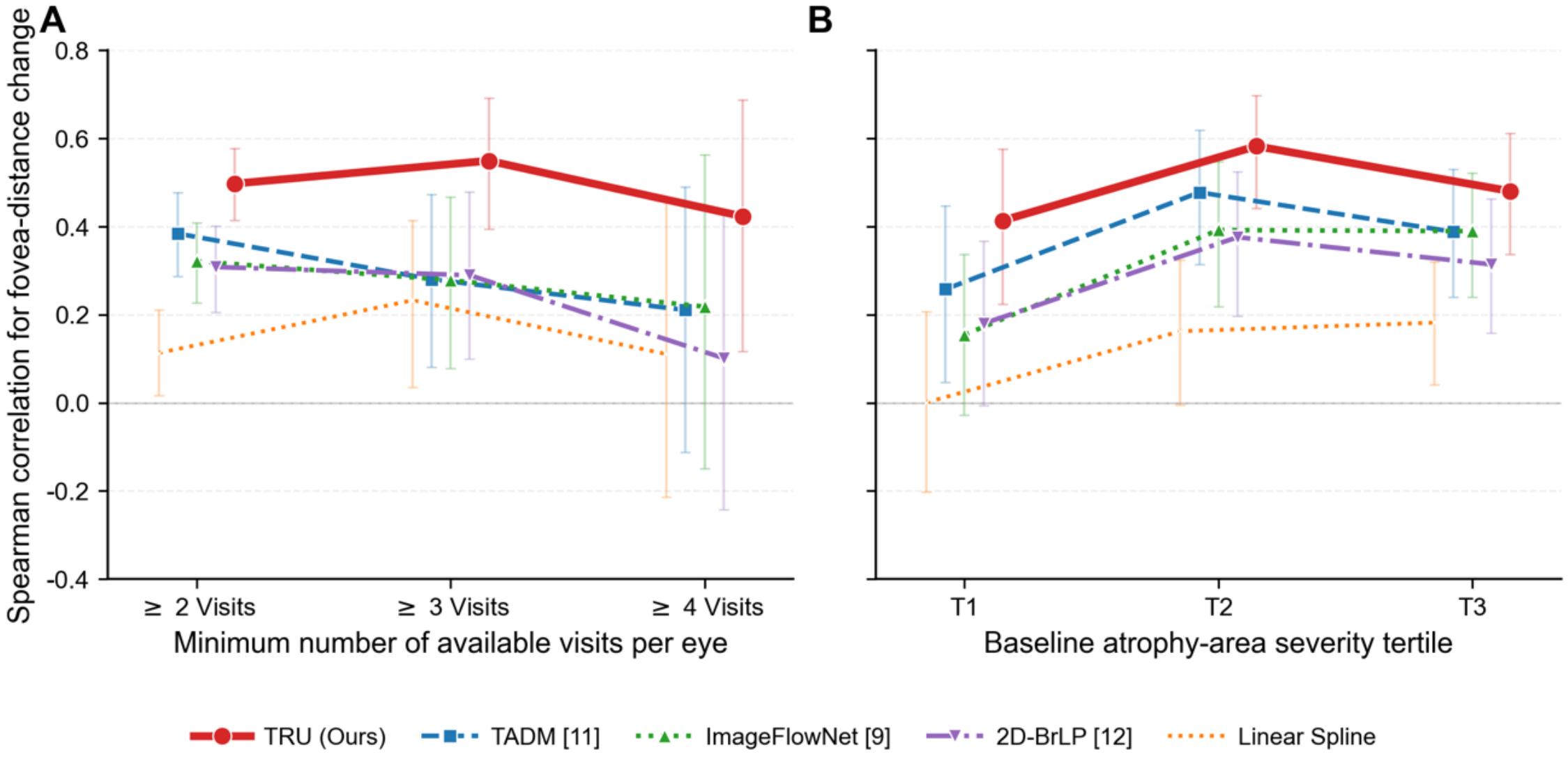

**Fig. 5.** Stratified robustness of fovea-distance-change prediction in the Optos FAF hold-out cohort. (A) Spearman correlation between predicted and observed fovea-to-atrophy distance change after restricting evaluation to eyes with increasing minimum numbers of available longitudinal visits. (B) Correlation coefficients for the same endpoint, fovea-to-atrophy distance change, stratified by tertiles of baseline atrophy burden (disease severity). Points denote patient-clustered bootstrap estimates; error bars denote 95% confidence intervals.

## IV. Discussion

This study makes two linked claims. First, model complexity for longitudinal prediction can be selected by task-specific measurements, and for slowly progressive FAF atrophy, that measurement favors a deterministic model. Second, an instantiation of the selected model class, TRU, achieved superior performance as compared to the evaluated, more complex models, and retained its advantage across disease and vendor shift, as well as across severity and history length strata.

Image series that are acquisition noise-dominated can still yield useful predictive information, because the elapsed-time sensitivity screen measures how much of the global image changes between visits, while a recoverable signal can still lie in where that change occurs. A prediction over a short time interval can have weak association with elapsed-time and still be spatially predictable, as was demonstrated in this study. The structure an aligned model recovers is the slow advance of an existing lesion edge and its shifting position relative to the fovea, which is exactly what the primary atrophy-to-fovea distance endpoint quantifies. A simplified prediction model that treats this outcome as a scalar value without using the image itself was anti-correlated with the change that followed (Section 3.2), confirming that the recoverable regularity does not lie in the endpoint's own scalar history but in the geometry of the image.

To determine the best approach for making those spatial predictions, the elapsed-time screen identifies a candidate regime, followed by checking whether stochastic seeds produced clinically resolvable variation and whether that variation improved the patient-level endpoint. In this task, empirical sample dispersion concentrated near a common prediction and remained below the endpoint measurement resolution for most eyes.

These results help explain why a compact model can perform well and why heavier stochastic machinery adds nothing in this regime. A network trained to denoise a corrupted version of the target image, then asked at inference to start from noise or from the most recent observation, is solving a problem it never saw in training, and in this study such a configuration underperformed even a simple no-progression prediction. Corrupting the conditioning input during training and starting inference from the same distribution removed that mismatch and accounted for most of the gap between stochastic and deterministic prediction. After alignment, the near future was highly similar to present anatomy, so repeated samples were almost identical and their spread was far too small to move the endpoint, which is why sampling bought nothing over a single deterministic pass. In this case, the deterministic variant is preferable in terms of parsimony, speed, and reproducibility.

Several limitation and caveats are worth noting. All cohorts, though spanning disease and vendor shift, come from a single institution. Multi-institutional data for generalization would be helpful. The study is intentionally based only on images, and does not link predicted endpoints to visual function, treatment decisions, or clinician behavior. Integrating those data are an area for future work and can help support clinical decision support uses of these techniques. Applying more sophisticated image preprocessing than what are currently implemented is also a planned future improvement, such as flat-field correction to remove vignetting across an image. Finally, the diagnostic governs model class, not architecture, and TRU is offered as one deterministic instance that other architectural designs may match or exceed.

## IV. Conclusion

Forecasting how disease progress from longitudinal imaging has been treated as a problem that calls for increasingly elaborate generative models. We show that the choice of model class can instead be guided by a diagnostic that measures whether changes between visits are dominated by progression signal or by technical noise, and tests whether stochastic sampling changes the clinical readout beyond measurement noise. For slow-progressing retinal atrophy prediction on FAF, this diagnostic selected the model class to be deterministic instead of stochastic. To test this choice, we built TRU, a deterministic model that predicts a future image from an irregular imaging history in a single forward pass. It outperformed all five classical and deep-learning benchmarks on image fidelity and on eye-level progression endpoints, both in held-out patients in two independent cohorts that differed in disease and imaging device. This work demonstrates how to measure whether stochasticity is required for a longitudinal imaging task, an approach that is not specific to the retinal imaging.

**Funding**

Research to Prevent Blindness Physician Scientist Award (J.C.); NIH R01 EY036222 (M.W.)

**Code and Data Availability Statement**

The source code for the Temporal Retinal U-Net (TRU), including training, inference, and evaluation scripts for all five conditioning configurations, is publicly available at *https://github.com/Liyincly01/TRU_longitudinal_retinal_image_prediction*. The raw imaging data of individual patients are not publicly available, but the fitted model weights are available at the link above.